\documentclass[final]{cvpr}

\usepackage{times}
\usepackage{epsfig}
\usepackage{graphicx}
\usepackage{amsmath}
\usepackage{amssymb}

\usepackage{multirow}
\usepackage{booktabs}
\usepackage{array}
\usepackage{colortbl}
\usepackage{color}
\usepackage{lipsum}
\usepackage{graphicx}
\usepackage{mathtools}
\usepackage{bm}
\usepackage{cite}
\usepackage{subfigure}
\usepackage{enumerate}
\usepackage{multicol}

\usepackage[pagebackref=true,breaklinks=true,colorlinks,bookmarks=false]{hyperref}

\def\cvprPaperID{91} % *** Enter the CVPR Paper ID here
\def\confYear{CVPR 2021}

\begin{document}

%%%%%%%%% TITLE
\title{ADNet: Attention-guided Deformable Convolutional Network for High Dynamic Range Imaging}

% \and

\author{%
Zhen Liu$^{1,2}$ \quad
Wenjie Lin$^1$ \quad
Xinpeng Li$^1$ \quad
Qing Rao$^1$ \quad
Ting Jiang$^1$ \quad \\
Mingyan Han$^1$ \quad
Haoqiang Fan$^1$\quad
Jian Sun$^1$\quad 
Shuaicheng Liu$^{3,1}$\thanks{Corresponding author.} \\
$^1$Megvii Technology \quad
$^2$Sichuan University \quad \\
$^3$University of Electronic Science and Technology of China
\\
\url{https://github.com/Pea-Shooter/ADNet}
%\texttt{\{luokunming,wangchuan,liushuaicheng,fhq,wangjue,sunjian\}@megvii.com}
}

\maketitle
\pagestyle{empty}  % no page number for the second and the later pages
\thispagestyle{empty} % no page number for the first page

%%%%%%%%% ABSTRACT
\begin{abstract}
   In this paper, we present an attention-guided deformable convolutional network for hand-held multi-frame high dynamic range (HDR) imaging, namely ADNet. This problem comprises two intractable challenges of how to handle saturation and noise properly and how to tackle misalignments caused by object motion or camera jittering. To address the former, we adopt a spatial attention module to adaptively select the most appropriate regions of various exposure low dynamic range (LDR) images for fusion. For the latter one, we propose to align the gamma-corrected images in the feature-level with a Pyramid, Cascading and Deformable (PCD) alignment module. The proposed ADNet shows state-of-the-art performance compared with previous methods, achieving a PSNR-$l$ of 39.4471 and a PSNR-$\mu$ of 37.6359 in NTIRE 2021 Multi-Frame HDR Challenge. 

\end{abstract}

%%%%%%%%% BODY TEXT
\section{Introduction}
High dynamic range imaging technique aims at recovering an HDR image from one or several LDR images. The former refers to single-frame HDR imaging~\cite{eilertsen2017hdr,lee2018deep,liu2020single,santos2020single}, while the latter refers to multi-frame HDR imaging~\cite{mertens2007exposure,ma2019deep,kalantari2017deep,wu2018deep,yan2019attention}. It has drawn much attention from low-level vision communities as traditional photography sensors cannot capture the actual dynamic range in nature scenes~\cite{nayar2000high,tumblin2005want}. Compared with single-frame HDR imaging, % multi-frame HDR imaging is more practical and promising because its input LDR images contain more information.
multi-frame HDR imaging is more practical and promising due to its informative bracket LDR inputs. If the LDR images are aligned perfectly, i.e., no object motion and camera jittering, the static images can be well fused~\cite{mertens2007exposure,ma2019deep}. When photographing with hand-held cameras, we need to handle the misalignments of various exposure LDR images first apart from image fusion. 

\begin{figure}[t]
   \includegraphics[width=1.0\linewidth]{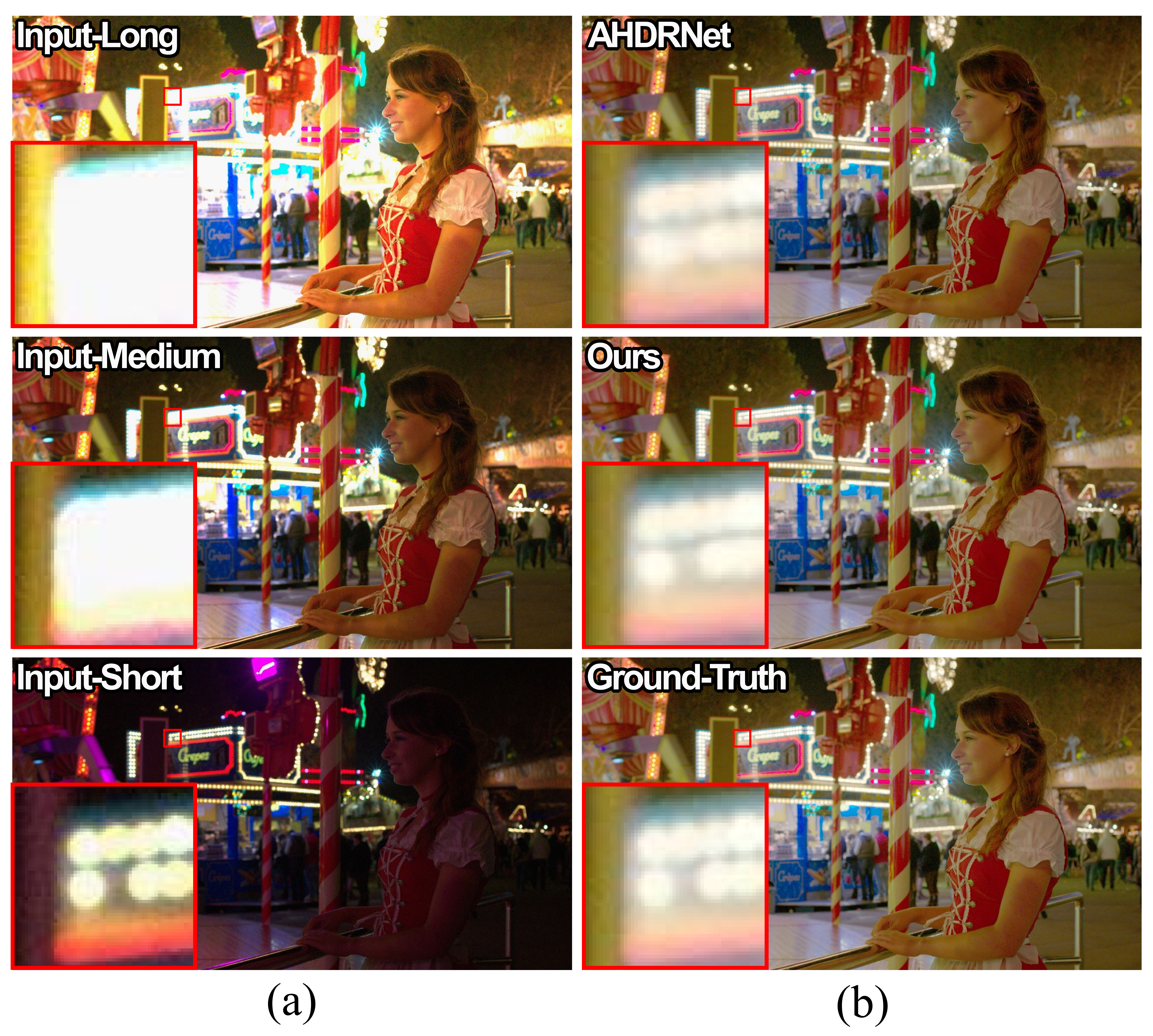}
   \caption{(a) LDR inputs with varying exposures. (b) Our result compared with the previous representative method AHDRNet. We show differences in the zoomed-in patches. Our result is free from noise and ghost artifacts while hallucinates more accurate details over the saturated areas. Zoom-in for a better comparison.}
   \label{fig:teaser}
   \end{figure}

Existing explicit image alignment methods mainly consists of three categories: global alignment with homography~\cite{hartley2003multiple}, middle-level alignment with meshflow~\cite{liu2016meshflow}, and pixel-level alignment with optical flow~\cite{baker2011database}. However, homography and meshflow cannot align foreground dynamic objects, and optical flow is erroneous in the presence of occlusion~\cite{luo2020upflow}. Several traditional methods are proposed to detect misaligned regions and then reject these pixels as outliers during the fusion process. However, these methods are often prone to introducing ghost artifacts as accurately identifying the dynamic objects is difficult.

Recently, several learning-based methods have been explored. Kalantari \emph{et al.} proposed the first deep convolutional neural network (CNN) for HDR imaging of dynamic scenes. They first aligned the LDR images with optical flow and then fused the aligned images by a CNN~\cite{kalantari2017deep}. However, optical flow is unreliable when occlusion and saturation occurs as mentioned above. Wu \emph{et al.} proposed the first non-flow-based approach, which performed homography alignment on LDR images before fed them into a CNN~\cite{wu2018deep}. Yan \emph{et al.} offered to handle motion by an attention module which achieved state-of-the-art results~\cite{yan2019attention} and later introduced a non-local neural network~\cite{yan2020deep}. All these methods processed the input LDR images and their gamma-corrected images uniformly, i.e., by simply concatenating them, resulting in blurry or ghost artifacts frequently.

In this paper, we propose an Attention-guided Deformable Convolutional Network (ADNet), a new pipeline to tackle such problems. Instead of directly concatenating the LDR images and their gamma-corrected images, we propose to process them with dual branches. Specifically, a spatial attention module is used for extracting the LDR images' attention features for better fusion, and a Pyramid, Cascading and Deformable (PCD) alignment module~\cite{wang2019edvr} is adopted to align the gamma-corrected images in the feature-level. Such design is motivated by the intuition that the images in the LDR domain help detect the noisy or saturated regions while the HDR counterparts help to detect misalignments~\cite{kalantari2017deep}. 

Fig.~\ref{fig:teaser} illustrates the results of our method and the state-of-the-art method AHDRNet~\cite{yan2019attention}. As can be seen, the results of AHDRNet fail to recover the details of saturated regions. In contrast, our method produces noise-free and ghost-free results while hallucinating more accurate contents in over saturated areas.
% contribution
Our main contribution can be summarized as follow: 
\begin{itemize}
   % \item We propose a novel dual branches pipeline for multi-frame HDR imaging of dynamic scenes. We process the LDR images with a spatial attention module and process the corresponding gamma-corrected images with a PCD align module.
   % \item The LDR and gamma-corrected images are often processed uniformly in the previous methods. Here, we process them separately, showing signiﬁcant improvements over previous opponents.
   \item We propose a novel dual-branch pipeline for multi-frame HDR imaging of dynamic scenes. Unlike previous methods that treat the LDR and gamma-corrected images uniformly, we process the LDR images with a spatial attention module and process the corresponding gamma-corrected images with a PCD alignment module.
   \item Existing learning-based methods use either optical flow based aligment or no explicit alignment. In this paper, we propose to align the dynamic frames with deformable alignment module, showing signiﬁcant improvements over previous opponents. To our best knowledge, this is the first application of deformable convolutions for multi-frame HDR imaging.
   \item Experimental results show that the proposed method achieves better results than the state-of-the-art methods, both quantitatively and qualitatively. Our approach also achieves the best results in NITRE 2021 Multi-Frame HDR Challenge.
 \end{itemize} 

%-------------------------------------------------------------------------
\section{Related Works}
We briefly summarize existing approaches into three categories: motion rejection methods, image registration methods, and learning-based methods.

Motion rejection methods perform global alignment upon LDR images first and then reject misaligned pixels before the image fusion. Gallo \emph{et al.} proposed to predict colors in various exposures directly and then compared them with original values to detect motion~\cite{gallo2009artifact}. Grosch \emph{et al.} calculated an error map according to the differences of alignment colors to reject misaligned pixels~\cite{grosch2006fast}. Pece \emph{et al.} detected the motion regions by computing the median threshold bitmap for input LDR images~\cite{pece2010bitmap}. Jacobs \emph{et al.} detected the misalignment regions by weighted intensity variance measurement~\cite{jacobs2008automatic}. Zhang \emph{et al.} computed a weight map of the LDR inputs in the gradient domain~\cite{zhang2011gradient}. Khan \emph{et al.} proposed to calculate the probability maps for pixels that belong to the background~\cite{khan2006ghost}. Oh \emph{et al.} also introduced rank minimization to detect ghost regions\cite{oh2014robust}. These methods often lead to unsatisfactory HDR effects as rejecting pixels will drop helpful information.

Image registration approaches register local image regions for fusion. Bogoni \emph{et al.} proposed to predict motion vectors using optical flow\cite{bogoni2000extending}. Hu \emph{et al.} proposed to perform image alignment in the transformation domain based on brightness and gradient consistencies\cite{hu2013hdr}. Kang \emph{et al.} converted LDR image intensities to the luminance domain according to exposure time and then calculated optical flow to compensate motion\cite{kang2003high}. Zimmer \emph{et al.} first registered the LDR images with optical flow and then recovered the HDR image~\cite{zimmer2011freehand}. Patch-based optimization is another type of image registration method besides optical flow. Sen \emph{et al.} proposed to align LDR images and reconstruct HDR image in a joint energy optimization process\cite{sen2012robust}. Jinno \emph{et al.} proposed to model the displacement with Markov random field\cite{jinno2008motion}. Image registration methods show better performance than motion rejection approaches. However, if large motion occurs, this approach introduces apparent artifacts.

Several deep learning approaches have been proposed recently\cite{kalantari2017deep,wu2018deep,yan2019multi, yan2019attention,eilertsen2017hdr,endo2017deep}. Single frame based methods recover HDR from a single LDR image. Eilertsen \emph{et al.} directly predicted HDR image from a single LDR input through a deep CNN\cite{eilertsen2017hdr}. Endo \emph{et al.} generated bracket LDR images from a single frame first and then fuse them to reconstruct HDR image\cite{endo2017deep}. Kalantari \emph{et al.} proposed the first deep multi-frame HDR imaging method of dynamic scenes. The LDR images were first aligned with optical flow and then blended by a fusion subnet\cite{kalantari2017deep}. Wu \emph{et al.} proposed the first non-flow-based deep framework. They performed a global registration on LDR inputs using homography and handled alignments and fusion through a UNet-based network~\cite{wu2018deep}. Yan \emph{et al.} proposed an attention-based deep CNN to control large motion, achieving the state-of-the-art eprformance~\cite{yan2019attention}. Our proposed method is built upon deep neural networks.

\begin{figure*}[t]
   \centering
   \includegraphics[width=1.0\linewidth]{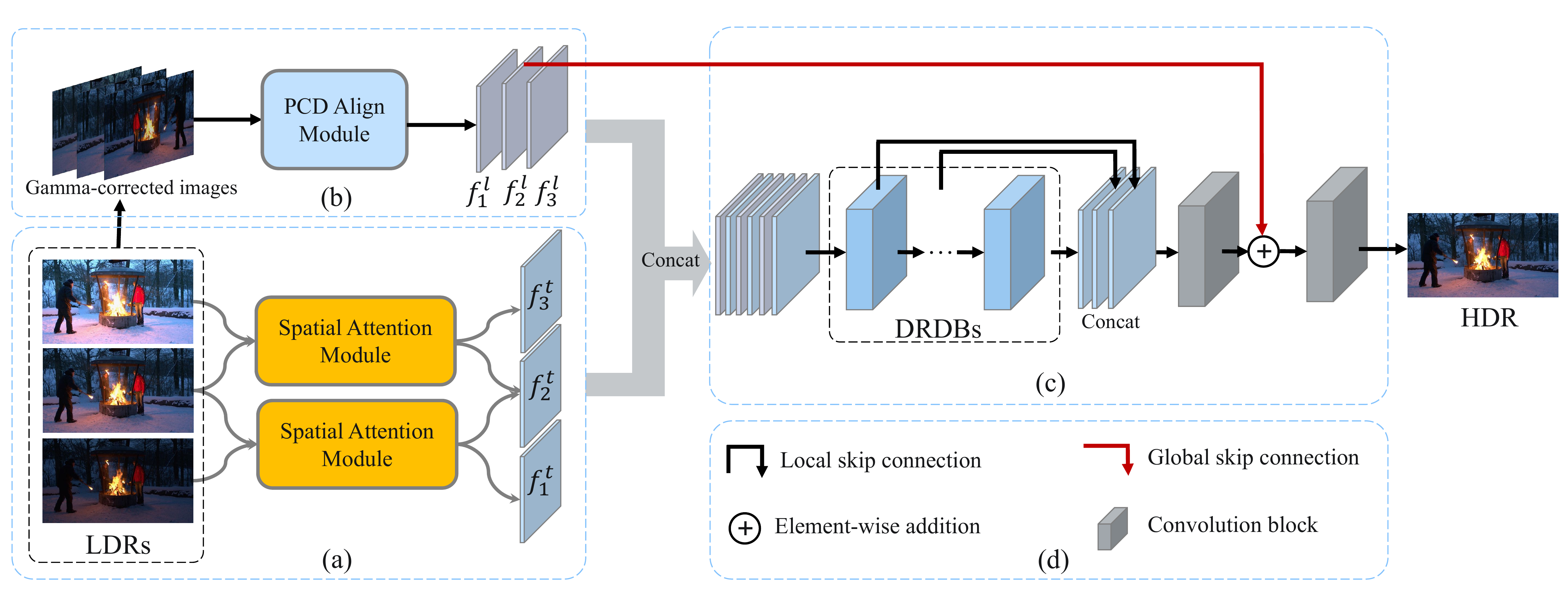}   
   \caption{The pipeline of our method. The network mainly consists of three components: (a) The input LDR images are ﬁrst fed into a spatial attention module to generate attention feature maps, and then (b) we adopt a PCD module to align the corresponding gamma-corrected images, (c) ﬁnally we employee several dilated residual dense blocks in the fusion subnet.}\label{fig:network-structure}
 \end{figure*}

%-------------------------------------------------------------------------
\section{Method}
The problem of multi-frame HDR imaging is to reconstruct an HDR image from several LDR images with various exposures. The middle frame of the LDR images is commonly selected as the reference image for motion alignment. In this paper, we consider 3 LDR images, i.e., $I_{i}, i=1, 2, 3$, as input and let the second LDR image $I_{2}$ be the reference image $I_{r}$. Existing learning-based methods~\cite{kalantari2017deep,wu2018deep,yan2019attention,yan2020deep} first map the input LDR images to the HDR domain using gamma correction and then concatenate them directly as the network input:

\begin{align}\label{eq:gamma_correction}
   \quad \check{I}_{i} = \frac{(I_{i})^\gamma}{t_{i}}, \quad i=1, 2, 3
\end{align}
where $t_{i}$ is the exposure time of $I_{i}$, $\gamma$ is the gamma correction parameter, and $\check{I}_{i}$ denotes the corresponding gamma-corrected images. 

% However, motivated by the intuition that the images in the LDR domain are helpful for detecting the noisy or saturated regions while the HDR counterparts help to detect misalignments, we propose a novel pipeline. Specifically, for the LDR images, we extract the attention feature maps with a spatial attention module $\mathcal{A}$. As for the gamma correction ones, we adpot a PCD align module $\mathcal{P}$ to handle dynamic objects or camera motions. Therefore, our proposed network $f$ can be defined as: 
Instead of concatenating them directly, we propose a novel dual-branch pipeline. Specifically, for the LDR images, we extract the attention feature maps with a spatial attention module $\mathcal{A}$. As for the gamma-corrected images, we adopt a PCD align module $\mathcal{P}$ to handle dynamic objects or camera motions. Therefore, our proposed network $f$ can be defined as: 
\begin{align}\label{eq:output}
   I^{\mathrm{H}} = f(\mathcal{A}(I_{i}), \mathcal{P}(\check{I}_{i}); \theta)
\end{align}
where $I^{H}$ denotes the reconstructed HDR image and $\theta$ denotes the network parameters.

\subsection{Network Structure}

We present the network structure of the proposed ADNet in Fig.~\ref{fig:network-structure}. The overall structure of ADNet mainly consists of three components: spatial attention module for LDR images (Fig.~\ref{fig:network-structure} (a)), PCD align module for gamma-corrected images (Fig.~\ref{fig:network-structure} (b)) and the fusion subnet for HDR reconstruction (Fig.~\ref{fig:network-structure} (c)). We first utilize a spatial attention module to extract the attention feature maps of three LDR images. Meanwhile, we use a PCD align module to align the gamma-corrected images in the feature level. Finally, we concatenate the attention features and the aligned gamma-corrected features in order of exposure time and feed them to the fusion subnet, generating the estimated HDR image.

\paragraph{Spatial Attention Module for LDR Images.} Given the LDR images $I_{i}, i=1, 2, 3$ with the shape of HxWx3, we first extract their LDR features by a single convolutional layer. For each non-reference LDR image $I_{i}, i\neq 2$, we concatenate the LDR feature with the feature of reference image as the input of the spatial attention module, generating the attention map with the range of 0-1. We then compute the element-wise multiplication of the LDR feature and its corresponding attention map to generate the spatial attention feature of each LDR image. The process can be formulated as 
\begin{align}\label{eq:attention_module}
   \quad f_{i}^{t} = \mathcal{A}(I_{i}, I_{r}), \quad i=1, 3
\end{align}
where $f_{2}^{t}$ is the reference feature. In this paper, we adopt the attention module as used in~\cite{yan2019attention}. The details are shown in Table~\ref{tab:attention_module}.

\begin{table}[h]
   \centering
   \caption{Details of the attention module used in our NTIRE 2021 HDR Challenge (Multi-Frame Track) submission.}
   \label{tab:attention_module} 
  \resizebox{1.0\linewidth}{!}{
    \begin{tabular}{
        p{3cm} % method
        >{\centering\arraybackslash}p{2.0cm} % PSNR-L
        >{\centering\arraybackslash}p{3.0cm} % PSNR-$\mu$ 
      }
      \toprule
         Name   & \# Out & Type\\
      \midrule
         Input & 128 & Input Feature Maps\\    
         Conv-Layer-1   & 128 & Conv 3x3\\ 
         ReLU-Layer-1   & 128 & ReLU Activation \\
         Conv-Layer-3   & 64 & Conv 3x3\\ 
         Sigmoid-Layer-4   & 64 & Sigmoid Activation \\    
         \bottomrule
      
     \end{tabular}
  }
 \end{table}

\paragraph{PCD Align Module for Gamma-corrected Images.} Performing alignment at the feature level is better than at the image level~\cite{chan2020understanding}. We also adopt this by employing deformable alignment~\cite{dai2017deformable}. Specifically, after applying gamma correction on the LDR images, we first extract the pyramid features using Stride convolutions and then perform deformable alignment with the reference feature on each scale of features, which is the same as the PCD module proposed by ~\cite{wang2019edvr}, i.e., 
\begin{align}\label{eq:pcd_align_module}
   \quad f_{i}^{l} = \mathcal{P}(\check{I}_{i}, \check{I}_{r}), \quad i=1, 3
\end{align}

\paragraph{Fusion Subnet for HDR Reconstruction.} The feature maps $f_{i}^{t}$ and $f_{i}^{l}$ are concatenated together as the input of the fusion subnet. Here, $f_{2}^{t}$ and $f_{2}^{l}$ denote the same reference features, which are not processed by the alignment or attention modules. The fusion subnet consists of several dilated residual dense blocks (DRDBs) which are also used in ~\cite{yan2019attention}. The usage of dilated convolution~\cite{yu2017dilated} increases the receptive field. We also use local skip connection and global skip connection for better training our model. The aforementioned network design generates ghost-free and noise-free HDR results, together with clearer image contents.

\subsection{Training Strategy} 

\begin{figure*}[t]
   \centering
   \includegraphics[width=1.0\linewidth]{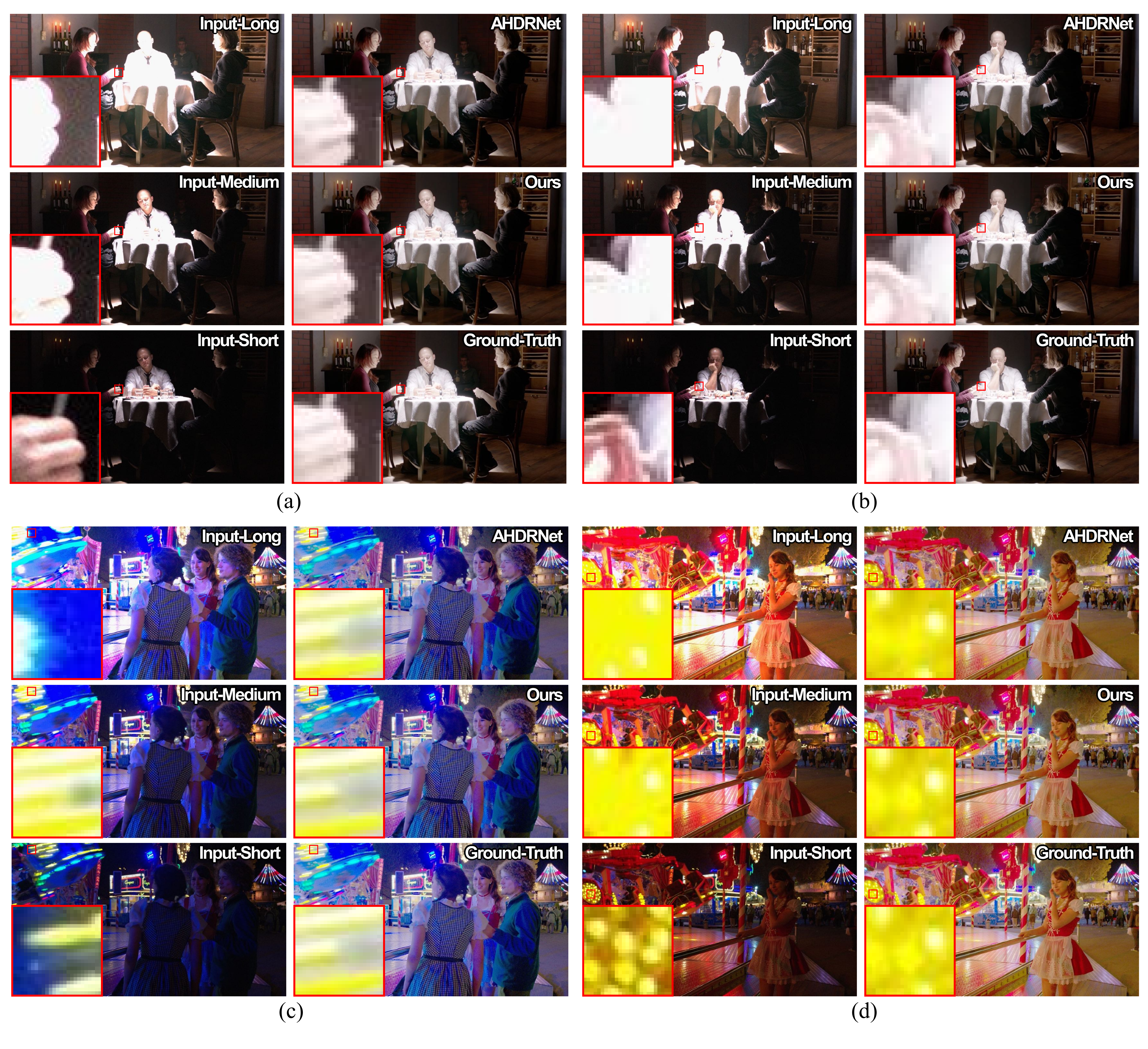}   
   \caption{Qualitative Comparison with existing state-of-the-art method AHDRNet. The proposed ADNet can not only reconstruct clearer image contents in motion boundaries (Fig.~\ref{fig:result_1} (a) and Fig.~\ref{fig:result_1} (b)) but also hallucinate more reasonable details in saturated regions (Fig.~\ref{fig:result_1} (c) and Fig.~\ref{fig:result_1} (d)).}
   \label{fig:result_1}
 \end{figure*}

\paragraph{Gamma Disturbance} Existing methods usually first linearize the non-linear LDR images using the camera response function (CRF)~\cite{debevec1997recovering} and then apply gamma correction(e.g., $\gamma=2.2$) on these linearized images to produce the input images~\cite{sen2012robust,kalantari2017deep}. However, as the CRF of each camera is not strictly the same, a fixed gamma value may not always be the most appropriate. In this paper, we propose a gamma disturbance strategy. Specifically, instead of maintaining a fixed gamma value all the time, we randomly apply a gamma value of $2.24\pm 0.1$ with the probability of 30$\%$. By adopting this strategy, The proposed method obtains about 0.1dB gain in terms of PSNR.

\paragraph{Loss Function} For multi-frame HDR imaging tasks, optimizing the network on the tonemapping domain is more effective than optimizing directly in the HDR domain as the HDR images are usually viewed after tonemapped~\cite{kalantari2017deep}. We also adopt such a strategy to train our ADNet. Given an estimated HDR image $I^{H}$ and the corresponding ground-truth HDR image $I^{GT}$, we first apply tonemapping onto them using the commonly used $\mu$-law: 
\begin{align}\label{eq:mu_law}
   \mathcal{T}(x)=\frac{\log (1+\mu x)}{1+\mu}.
\end{align}
We set $\mu=5000$ in this paper. Then we compute the $l_{1}$ error of the tonemapped images as our loss function, i.e., 
\begin{align}\label{eq:loss}
\mathcal{L}=\left\|\mathcal{T}\left(I^{\mathrm{GT}}\right)-\mathcal{T}\left(I^{H}\right)\right\|_{1}
\end{align}

%-------------------------------------------------------------------------
\section{Experiments}

\subsection{Dataset and Implementation Details}

We train and evaluate our method on the dataset provided by the NTIRE2021 Multi-Frame HDR Challenge~\cite{perez2021ntire}. It contains 1463 valid scenes in total as we exclude 31 incomplete scenes. We select 100 scenes as the validation set and keep the remaining as the training set. Each scene consists of three LDR images with various exposures and their corresponding HDR ground truth.

\begin{figure*}[t]
   \centering
   \includegraphics[width=1.0\linewidth]{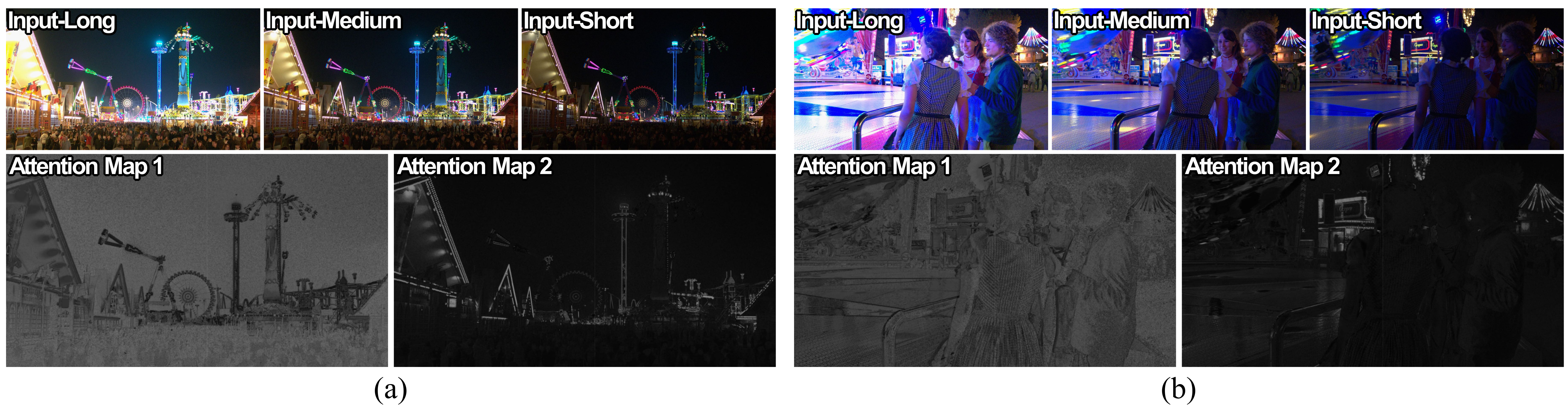}   
   \caption{Visualization of attention maps. The attention maps adaptively select the most appropriate regions of LDR images to fusion. In each scene of Fig.~\ref{fig:attention-map-vis} (a) and Fig.~\ref{fig:attention-map-vis} (b), the first attention map tends to suppress the over-exposure regions and vice-versa of the second one. Zoom-in for a better view.}\label{fig:attention-map-vis}
 \end{figure*}

During the training stage, we first crop the input LDR images to 256x256-sized patches with a stride of 128. The network is optimized by an Adam optimizer~\cite{kingma2014adam} with initial learning rate of 1e-4 and decay rate of 0.1, we set $\beta_{1}=0.9$, $\beta_{2}=0.999$ and $\epsilon=10^{-8}$. Our experiments are implemented in PyTorch and trained on 8 NVIDIA 2080Ti GPUs with batch size of 16. We train the model from scratch for 200 epochs, and the whole training costs about three days. We select the best model using the PSNR-$\mu$ score calculated on our validation set when the training reaches plateaus.

The entire testing process is conducted on a single 2080Ti GPU. Constrained by the limited GPU memory, we split each test image into size 1060x1000 and 1060x900 for testing and then concatenate them to the full size 1060x1900. We compute the PSNR-$l$ and PSNR-$\mu$ scores as the testing metrics where `-$l$' and `-$\mu$' means the ones computed in the linear domain and the tonemapped domain, respectively.

\begin{table}[h]
   \centering
   \caption{Quantitive comparison with AHDRNet. The PSNR-$l$ and PSNR-$\mu$ refer to the PSNR scores computed in the linear domain and tonemapped domain. The `TTA' means testing-time augmentation.}
   \label{tab:quantitive_results} 
	\resizebox{1.0\linewidth}{!}{
		\begin{tabular}{
				p{1.5cm} % method
				>{\centering\arraybackslash}p{1.5cm} % method
				% >{\centering\arraybackslash}p{1.4cm}
				>{\centering\arraybackslash}p{2.0cm} % PSNR-L
				>{\centering\arraybackslash}p{2.0cm} % PSNR-$\mu$ 
				% >{\centering\arraybackslash}p{1.0cm} % SSIM-$\mu$
				% >{\centering\arraybackslash}p{1.0cm} % SSIM-L  
				% >{\centering\arraybackslash}p{1.6cm} % HDR-VDP-2 
			}
			\toprule
			\multicolumn{2}{l}{}  & PSNR-$l$   & PSNR-$\mu$   \\ 
			\midrule
			\multirow{2}{*}{AHDRNet}    & w/o. TTA & 38.6737   & 36.7068   \\
			                        & w/ TTA    & 39.0577   & 36.8073   \\
			\midrule
			
			\multirow{2}{*}{Ours}    & w/o. TTA & 39.3398   & 37.2068  \\
			& w/ TTA   & \textbf{39.9167} & \textbf{37.3548}   \\
         \bottomrule
      
	   \end{tabular}
	}
\end{table}

\subsection{Results and Analysis}

To demonstrate the superiority of our proposed ADNet, we compare it with the existing state-of-the-art method AHDRNet~\cite{yan2019attention}, both quantitively and qualitatively. For fair comparisons, we retrain AHDRNet in the challenge data with the same settings as ours and report the PSNR computed in the linear domain and tonemapped domain, i.e., PSNR-$l$ and PSNR-$\mu$. As listed in Table~\ref{tab:quantitive_results}, the proposed ADNet produces higher scores among all the calculated metrics, outperforming the AHDRNet by 0.85dB in terms of PSNR-$l$ and 0.55dB in terms of PSNR-$\mu$. % Note that the `TTA' means testing-time augmentation, which can be seen as a self-ensemble strategy.  % To boost performance, we apply the testing-time augmentation (TTA) strategy, which can be seen as a self-ensemble strategy. Specifically, 
The `TTA' means testing-time augmentation, which can be seen as a self-ensemble strategy. In this paper, we apply a `4x' version where the final result is averaged from four versions of the input images, i.e., the original, the permuted, the vertically flipped, and the horizontally flipped images.

We also report the qualitative results compared with AHDRNet. As illustrated in Fig.~\ref{fig:result_1}, The first two scenes (Fig.~\ref{fig:result_1} (a) and Fig.~\ref{fig:result_1} (b)) contain subtle motion upon the lady's hands, and meanwhile, the middle frame and the long-time exposure frame encounter saturation in these regions. The result of AHDRNet produces undesirable details and blurry contents while ours are clearer and more precise in motion boundaries. Fig.~\ref{fig:result_1} (c) is more intractable as the motion magnitude gets larger upon the saturated hobbyhorse. The scene of Fig.~\ref{fig:result_1} (d) encounters more severe saturation regions. As seen, the results of AHDRNet fail to recover the corresponding areas while our method can hallucinate reasonable details and is free of ghost artifacts and noises.

% Analysis
We attribute the de-ghosting and HDR reconstruction ability of our method to the new network design. On the one hand, the spatial attention module performed on the LDR images can adaptively select proper image regions for fusion, i.e., the light regions in the short-time exposure frame and the dark region in the long-time exposure frame. To futher verify this, we visualize the attention maps by averaging them along the channel dimension. As shown in Fig.~\ref{fig:attention-map-vis}, the attention maps suppress the over-/under-exposure regions and focus on the well-exposed areas. On the other hand, the PCD align module handles the gamma-corrected images which contain camera motions or dynamic objects and aligns them more accurately. We also verify the effective network design through ablation studies.

\subsection{Ablation Study}

To verify the effectiveness of the proposed ADNet, we conduct ablation studies of the network architecture and analyze the results. It should be noted that all the ablation studies are compared in our validation set as the test set is currently unavailable. We compare our method with three variants as follow:
\begin{itemize}
   \item \textbf{Baseline}. Our method takes the AHDRNet~\cite{yan2019attention} as our baseline, which contains an attention network and a merging network.
   \item \textbf{Variant 1}. This variant replaces the attention module in the baseline as a vanilla deformable alignment module. Specifically, we apply deformable alignment only on the original scale features instead of pyramid features. 
   \item \textbf{Variant 2}. This variant adopts a PCD alignment module instead of a single scale one as used in the first variant.
   \item \textbf{Ours}. The entire network architecture of the proposed dual-branch ADNet, which contains a spatial attention module and a PCD align module.
 \end{itemize} 

 \paragraph{PCD Align V.S. Vanilla Deformable Align} As shown in Table~\ref{tab:ablation_study}, the first variant only employs the vanilla deformable alignment with one single scale while the second variant adopts a PCD align module as used in~\cite{wang2019edvr}. Compared with the baseline model, the deformable alignment shows better performance. With the usage of pyramid features alignment, the second variant has a 1.2dB gain of PSNR-$l$. The main reason can be concluded as that the PCD module enriches the feature representation ability, and aligns the features across multi-level features can handle more complex motions.

 \begin{table}[t]
   \caption{Ablation studies on the network architecture.}
   \label{tab:ablation_study}
   \centering
   \resizebox{1.0\linewidth}{!}{
     \begin{tabular}{
     l%p{cm}
     >{\centering\arraybackslash}p{2.8cm}
     >{\centering\arraybackslash}p{2.8cm}}
     \toprule
                     & PSNR-$l$  & PSNR-$\mu$  \\
     \midrule
     Baseline          & 41.7593              & 34.6083 \\

     \midrule
     Variant 1	         & 42.0612	& 34.6113 	\\
     Variant 2	          & 43.2681	& 34.7316               \\
     \midrule
     Ours            & \textbf{43.6218}          & \textbf{34.8606}\\
     \bottomrule
     \end{tabular}
   }
 \end{table}%

 \paragraph{Dual Branches V.S. Single Branch} We also conduct experiments to explore the effectiveness of the dual branches design. As shown in Table~\ref{tab:ablation_study}, the baseline model and its two variants are designed as a single branch, i.e., concatenating the LDR images and the gamma-corrected images directly. The results show that the dual branches design, which treats the LDR images and their gamma-corrected images separately, is more effective, especially when compared with the baseline model AHDRNet.

% \begin{table}[t]
%    \caption{Ablation studies.}
%    \label{tab:ablation_study}
%    \centering
%    \resizebox{1.0\linewidth}{!}{
%      \begin{tabular}{
%      l%p{cm}
%      >{\centering\arraybackslash}p{1.4cm}
%      >{\centering\arraybackslash}p{1.4cm}
%      >{\centering\arraybackslash}p{1.4cm}
%      >{\centering\arraybackslash}p{1.4cm}}
%      \toprule
%                      & PSNR-$l$  & PSNR-$\mu$    & SSIM-$l$  & SSIM-$\mu$    \\
%      \midrule
%      Baseline          & 43.1167              & 40.8782              & 0.9855    & 0.9841       \\

%      \midrule
%      Variant 1	         & 41.8864	& 40.8764     & 0.9846              & 0.9832	\\
%      Variant 2	                 & 43.4612	& 40.9769	  & 0.9896	            & 0.9865	 \\
%      \midrule
%      Ours            & \textbf{43.8907}          & \textbf{41.4164}           & \textbf{0.9908}  & \textbf{0.9871}  \\
%      \bottomrule
%      \end{tabular}
%    }
%  \end{table}%

%------------------------------------------------------------------------
\section{Conclusion}
We have presented ADNet, an attention-guided deformable convolutional network for multi-frame HDR imaging. A dual-branch pipeline is proposed where we handle the LDR images with a spatial attention module, and tackle misalignments with a PCD align module. Experimental results show that the proposed method can achieve state-of-the-art performance and reconstruct noise-free and ghost-free HDR images. Code used in this work will be publicly available upon publication.

\section*{Acknowledgement}
This work was supported by the National Natural Science Foundation of China (NSFC) under Grants No.61872067 and No.61720106004.

{\small
\bibliographystyle{ieee_fullname}
\bibliography{egbib}
}

\end{document}